%% file: benchmark.tex
\RequirePackage[loading]{tracefnt}

\documentclass{article}

\usepackage{times}
\usepackage{graphicx} 
\usepackage{subfigure} 
\usepackage{dblfloatfix}
\usepackage{natbib}

\usepackage{caption}
\newcommand{\ignore}[2]{\hspace{0in}#2}

\usepackage{algorithm}
\usepackage{algorithmic}

\usepackage{hyperref}

\usepackage[accepted]{icml2016}
\usepackage{verbatim}
\usepackage{float}
\usepackage{rotating}
\usepackage{siunitx}
\usepackage{array}

\usepackage{colortbl,etoolbox,xcolor}
\robustify\bfseries \include{defs}

\icmltitlerunning{Benchmarking Deep Reinforcement Learning for Continuous Control}

\begin{document} 

\twocolumn[
\icmltitle{Benchmarking Deep Reinforcement Learning for Continuous Control}

\icmlauthor{Yan Duan$^\dag$}{rockyduan@eecs.berkeley.edu}
\icmlauthor{Xi Chen$^\dag$}{c.xi@eecs.berkeley.edu}
\icmlauthor{Rein Houthooft${^\dag}^\ddag$}{rein.houthooft@ugent.be}
\icmlauthor{John Schulman${^\dag}^\S$}{joschu@eecs.berkeley.edu}
\icmlauthor{Pieter Abbeel$^\dag$}{pabbeel@cs.berkeley.edu}
\icmladdress{$^\dag$ University of California, Berkeley, Department of Electrical Engineering and Computer Sciences\\
$^\ddag$ Ghent University - iMinds, Department of Information Technology\\
$\S$ OpenAI
}
\icmlkeywords{Reinforcement Learning, Deep Learning, Benchmark, Robotics}

\vskip 0.3in
]

\renewcommand*{\UrlFont}{\normalsize}
\begin{abstract}
Recently, researchers have made significant progress combining the advances in deep learning for learning feature representations with reinforcement learning. Some notable examples include training agents to play Atari games based on raw pixel data and to acquire advanced manipulation skills using raw sensory inputs. However, it has been difficult to quantify progress in the domain of continuous control due to the lack of a commonly adopted benchmark. In this work, we present a benchmark suite of continuous control tasks, including classic tasks like cart-pole swing-up, tasks with very high state and action dimensionality such as 3D humanoid locomotion, tasks with partial observations, and tasks with hierarchical structure. We report novel findings based on the systematic evaluation of a range of implemented reinforcement learning algorithms. Both the benchmark and reference implementations are released at \url{https://github.com/rllab/rllab} in order to facilitate experimental reproducibility and to encourage adoption by other researchers.
\end{abstract}

\section{Introduction}

\renewcommand*{\UrlFont}{\normalsize}

Reinforcement learning addresses the problem of how agents should learn to take actions to maximize cumulative reward through interactions with the environment. The traditional approach for reinforcement learning algorithms requires carefully chosen feature representations, which are usually hand-engineered. 
Recently, significant progress has been made by combining advances in deep learning for learning feature representations \cite{Krizhevsky12, Hinton12} with reinforcement learning, tracing back to much earlier work of \citet{Tesauro95TDGammon} and \citet{Bertsekas95Neuro}. Notable examples are training agents to play Atari games based on raw pixels \cite{NIPS2014_5421, Mnih15, Schulman15TRPO}
and to acquire advanced manipulation skills using raw sensory inputs \cite{Levine15, Lillicrap15, Watter15E2C}. Impressive results have also been obtained in training deep neural network policies for 3D locomotion and manipulation tasks \cite{Schulman15TRPO, Schulman15GAE, Heess15}.

Along with this recent progress, the Arcade Learning Environment (ALE) \cite{Bellemare13ALE} has become a popular benchmark for evaluating algorithms designed for tasks with high-dimensional state inputs and discrete actions. However, these algorithms do not always generalize straightforwardly to tasks with continuous actions, leading to a gap in our understanding. For instance, algorithms based on Q-learning quickly become infeasible when naive discretization of the action space is performed, due to the curse of dimensionality \cite{Bellman57, Lillicrap15}. In the continuous control domain, where actions are continuous and often high-dimensional, we argue that the existing control benchmarks fail to provide a comprehensive set of challenging problems (see Section~\ref{section:related_work} for a review of existing benchmarks).
Benchmarks have played a significant role in other areas such as computer vision and speech recognition. Examples include MNIST \cite{lecun1998mnist}, Caltech101 \cite{fei2006one}, CIFAR \cite{krizhevsky2009learning}, ImageNet \cite{Deng09ImageNet}, PASCAL VOC \cite{everingham2010pascal}, BSDS500 \cite{MartinFTM01}, SWITCHBOARD \cite{sb}, TIMIT \cite{garofolo1993darpa}, Aurora \cite{hirsch2000aurora}, and VoiceSearch \cite{Dong07VoiceSearch}. The lack of a standardized and challenging testbed for reinforcement learning and continuous control makes it difficult to quantify scientific progress. Systematic evaluation and comparison will not only further our understanding of the strengths of existing algorithms, but also reveal their limitations and suggest directions for future research.

We attempt to address this problem and present a benchmark consisting of 31 continuous control tasks. These tasks range from simple tasks, such as cart-pole balancing, to challenging tasks such as high-DOF locomotion, tasks with partial observations, and hierarchically structured tasks. Furthermore, a range of reinforcement learning algorithms are implemented on which we report novel findings based on a systematic evaluation of their effectiveness in training deep neural network policies. The benchmark and reference implementations are available at \url{https://github.com/rllab/rllab}, allowing for the development, implementation, and evaluation of new algorithms and tasks. 


\section{Preliminaries}
\label{section:preliminaries}

In this section, we define the notation used in subsequent sections.

The implemented tasks conform to the standard interface of a finite-horizon discounted Markov decision process (MDP), defined by the tuple $(\cS, \cA, P, r, \rho_0, \gamma, T)$, where $\cS$ is a (possibly infinite) set of states, $\cA$ is a set of actions, $P: \cS \times \cA \times \cS \rightarrow \bR_{\geq 0}$ is the transition probability distribution, $r: \cS \times \cA \rightarrow \bR$  is the reward function, $\rho_0: \cS \to \bR_{\geq 0}$ is the initial state distribution, $\gamma \in (0, 1]$ is the discount factor, and $T$ is the horizon.

For partially observable tasks, which conform to the interface of a partially observable Markov decision process (POMDP), two more components are required, namely $\Omega$, a set of observations, and $\cO: \cS \times \Omega \to \bR_{\geq 0}$, the observation probability distribution.

Most of our implemented algorithms optimize a stochastic policy $\pi_\theta: \cS \times \cA \rightarrow \bR_{\geq 0}$. Let $\eta(\pi)$ denote its expected discounted reward: 
$ \eta(\pi) = \bE_{\tau}\left[ \sum_{t=0}^T \gamma^t r(s_t, a_t) \right]$, where $\tau = (s_0, a_0, \ldots)$ denotes the whole trajectory, $\displaystyle s_0 \sim \rho_0(s_0)$, $a_t \sim \pi(a_t|s_t)$, and $s_{t+1} \sim P(s_{t+1} | s_t, a_t)$.

For deterministic policies, we use the notation $\mu_\theta: \cS \rightarrow \cA$ to denote the policy instead. The objective for it has the same form as above, except that now we have $a_t = \mu(s_t)$.

\section{Tasks}

The tasks in the presented benchmark can be divided into four categories: basic tasks, locomotion tasks, partially observable tasks, and hierarchical tasks.
We briefly describe them in this section.
More detailed specifications are given in the supplementary materials and in the source code.

We choose to implement all tasks using physics simulators rather than symbolic equations, since the former approach is less error-prone and permits easy modification of each task. Tasks with simple dynamics are implemented using Box2D \cite{Box2D}, an open-source, freely available 2D physics simulator. Tasks with more complicated dynamics, such as locomotion, are implemented using MuJoCo \cite{MuJoCo}, a 3D physics simulator with better modeling of contacts.


\subsection{Basic Tasks}
\label{sec:basic_tasks}




We implement five basic tasks that have been widely analyzed in reinforcement learning and control literature: Cart-Pole Balancing \cite{stephenson1908xx, donaldson1960error, widrow1964pattern, boxes}, Cart-Pole Swing Up \cite{815604, doya2000reinforcement}, Mountain Car \cite{Moore90MountainCar}, Acrobot Swing Up \cite{dejong1994swinging,Murray:M91/46,doya2000reinforcement}, and Double Inverted Pendulum Balancing \cite{Furuta78}. These relatively low-dimensional tasks provide quick evaluations and comparisons of RL algorithms.

\subsection{Locomotion Tasks}
\label{sec:locomotion_tasks}

In this category, we implement six locomotion tasks of varying dynamics and difficulty: Swimmer \cite{PurcellSwimmer, coulom2002reinforcement, levine2013guided, Schulman15TRPO}, Hopper \cite{murthy19843d, erez2011infinite, levine2013guided, Schulman15TRPO}, Walker \cite{raibert1991animation, erez2011infinite, levine2013guided, Schulman15TRPO}, Half-Cheetah \cite{wawrzynski2007learning, Heess15}, Ant \cite{Schulman15GAE}, Simple Humanoid \cite{tassa2012synthesis, Schulman15GAE}, and Full Humanoid \cite{tassa2012synthesis}.
The goal for all the tasks is to move forward as quickly as possible. These tasks are more challenging than the basic tasks due to high degrees of freedom. In addition, a great amount of exploration is needed to learn to move forward without getting stuck at local optima. Since we penalize for excessive controls as well as falling over, during the initial stage of learning, when the robot is not yet able to move forward for a sufficient distance without falling, apparent local optima exist including staying at the origin or diving forward slowly.


\begin{figure}[!h]
\subfigure[]{
\includegraphics[height=53px]{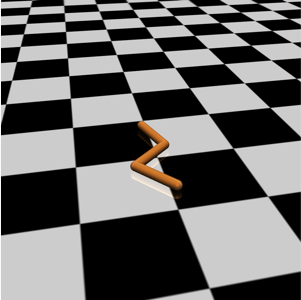}\label{fig:swimmer}}
\subfigure[]{
\includegraphics[height=53px]{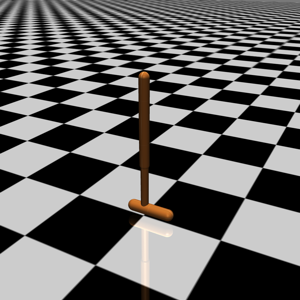}\label{fig:hopper}}
\subfigure[]{
\includegraphics[height=53px]{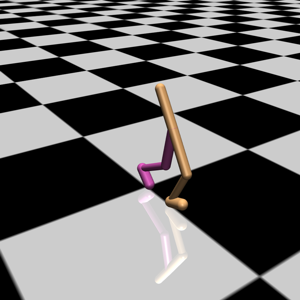}\label{fig:walker}}
\subfigure[]{
\includegraphics[height=53px]{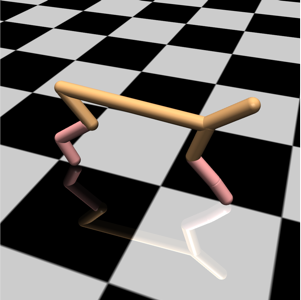}\label{fig:half_cheetah}}
\subfigure[]{
\includegraphics[height=53px]{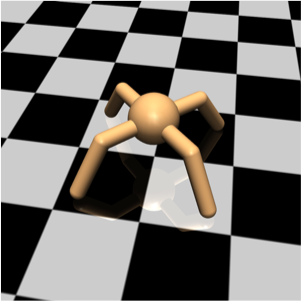}\label{fig:ant}}
\subfigure[]{
\includegraphics[height=53px]{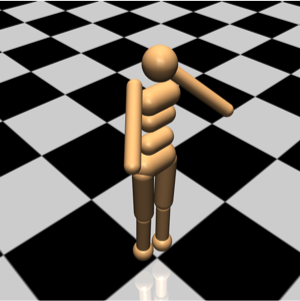}\label{fig:simple_humanoid}}
\subfigure[]{
\includegraphics[height=53px]{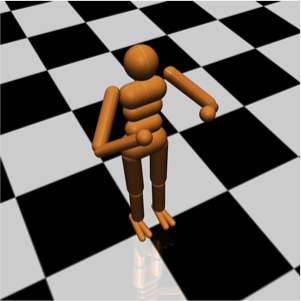}\label{fig:humanoid}}
\caption{Illustration of locomotion tasks: \subref{fig:swimmer} Swimmer; \subref{fig:hopper} Hopper; \subref{fig:walker} Walker; \subref{fig:half_cheetah} Half-Cheetah; \subref{fig:ant} Ant; \subref{fig:simple_humanoid} Simple Humanoid; and \subref{fig:humanoid} Full Humanoid.}\label{fig:plots_locomotion_tasks}
\end{figure}

\subsection{Partially Observable Tasks}

In real-life situations, agents are often not endowed with perfect state information. This can be due to sensor noise, sensor occlusions, or even sensor limitations that result in partial observations. To evaluate algorithms in more realistic settings, we implement three variations of partially observable tasks for each of the five basic tasks described in Section~\ref{sec:basic_tasks}, leading to a total of $15$ additional tasks. These variations are described below.

{\bf Limited Sensors}: For this variation, we restrict the observations to only provide positional information (including joint angles), excluding velocities. An agent now has to learn to infer velocity information in order to recover the full state. Similar tasks have been explored in \citet{gomez19982, schafer2005solving, heess2015memory, wierstra2007solving}.

{\bf Noisy Observations and Delayed Actions}: In this case, sensor noise is simulated through the addition of Gaussian noise to the observations. We also introduce a time delay between taking an action and the action being in effect, accounting for physical latencies \cite{hester2013open}. Agents now need to learn to integrate both past observations and past actions to infer the current state. Similar tasks have been proposed in \citet{bakker2001reinforcement}.

{\bf System Identification}: For this category, the underlying physical model parameters are varied across different episodes \cite{szita2003varepsilon}. The agents must learn to generalize across different models, as well as to infer the model parameters from its observation and action history. 

\subsection{Hierarchical Tasks}
\label{subsection:hierarchical_tasks}

Many real-world tasks exhibit hierarchical structure, where higher level decisions can reuse lower level skills \cite{parr1998reinforcement, sutton1999between, Dietterich:2000}. For instance, robots can reuse locomotion skills when exploring the environment. We propose several tasks where both low-level motor controls and high-level decisions are needed. These two components each operates on a different time scale and calls for a natural hierarchy in order to efficiently learn the task.


\begin{figure}[!h]
\centering
\subfigure[]{
\includegraphics[height=90px]{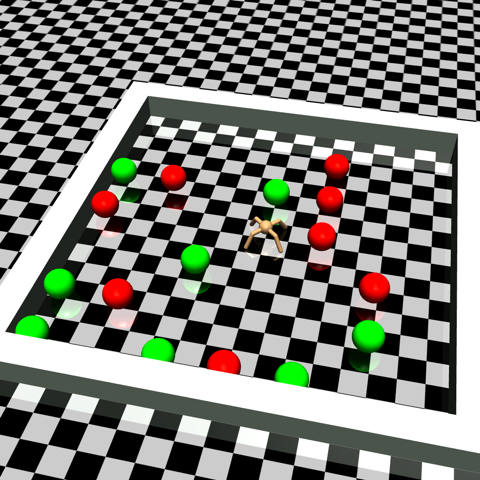}\label{fig:ant_gather}}
\subfigure[]{
\includegraphics[height=90px]{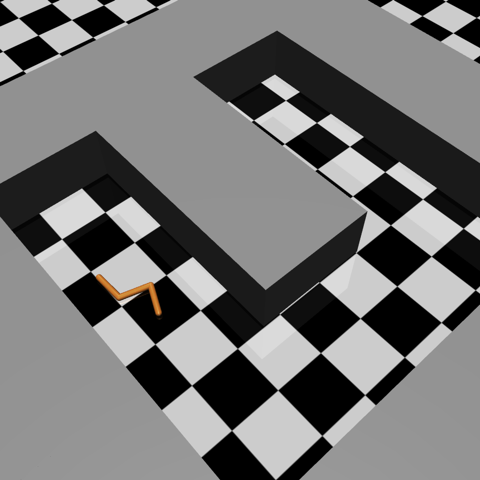}\label{fig:swimmer_maze}}
\caption{Illustration of hierarchical tasks: \subref{fig:ant_gather} Locomotion + Food Collection; and \subref{fig:swimmer_maze} Locomotion + Maze.}\label{fig:plots_hierarchical_tasks}
\end{figure}


    
{\bf Locomotion + Food Collection}: For this task, the agent needs to learn to control either the swimmer or the ant robot to collect food and avoid bombs in a finite region. The agent receives range sensor readings about nearby food and bomb units. It is given a positive reward when it reaches a food unit, or a negative reward when it reaches a bomb.

{\bf Locomotion + Maze}: For this task, the agent needs to learn to control either the swimmer or the ant robot  to reach a goal position in a fixed maze. The agent receives range sensor readings about nearby obstacles as well as its goal (when visible). A positive reward is given only when the robot reaches the goal region.



\section{Algorithms}

In this section, we briefly summarize the algorithms implemented in our benchmark, and note any modifications made to apply them to general parametrized policies. We implement a range of gradient-based policy search methods, as well as two gradient-free methods for comparison with the gradient-based approaches.


\subsection{Batch Algorithms}
Most of the implemented algorithms are batch algorithms. At each iteration, $N$ trajectories $\{ \tau_i\}_{i=1}^N$ are generated, where
$\tau_i =\{ (s_t^i, a_t^i, r_t^i)\}_{t=0}^T$ contains data collected along the $i$th trajectory. For on-policy gradient-based methods, all the trajectories are sampled under the current policy. For gradient-free methods, they are sampled under perturbed versions of the current policy.
    


{\bf REINFORCE} \cite{Williams92VPG}: This algorithm estimates the gradient of expected return $\nabla_\theta \eta(\pi_\theta)$ using the likelihood ratio trick:
\[\widehat{\nabla_\theta \eta(\pi_\theta)} = \frac{1}{NT} \sum_{i=1}^N \sum_{t=0}^T \nabla_\theta \log \pi(a_t^i|s_t^i; \theta) (R_t^i - b_t^i),\]
where $R_t^i = \sum_{t'=t}^{T} \gamma^{t'-t} r_{t'}^i$ and $b_t^i$ is a baseline that only depends on the state $s_t^i$ to reduce variance. Hereafter, an ascent step is taken in the direction of the estimated gradient. This process continues until $\theta_k$ converges.

{\bf Truncated Natural Policy Gradient (TNPG)} \cite{Kakade02NPG, Peters03PG, Bagnell03Cov, Schulman15TRPO}: Natural Policy Gradient improves upon REINFORCE by computing an ascent direction that approximately ensures a small change in the policy distribution. This direction is derived to be $I(\theta)^{-1} \nabla_\theta \eta(\pi_\theta)$, where $I(\theta)$ is the Fisher information matrix (FIM). We use the step size suggested by \citet{peters2008reinforcement}: $\alpha = \sqrt{\delta_{\text{KL}} \left(\nabla_\theta \eta(\pi_\theta)^T I(\theta)^{-1} \nabla_\theta \eta(\pi_\theta)\right)^{-1}}$. Finally, we replace $\nabla_\theta \eta(\pi_\theta)$ and $I(\theta)$ by their empirical estimates.


For neural network policies with tens of thousands of parameters or more, generic Natural Policy Gradient incurs prohibitive computation cost by forming and inverting the empirical FIM. Instead, we study Truncated Natural Policy Gradient (TNPG) in this paper, which computes the natural gradient direction without explicitly forming the matrix inverse, using a conjugate gradient algorithm that only requires computing $I(\theta)v$ for arbitrary vector $v$. TNPG makes it practical to apply natural gradient in policy search setting with high-dimensional parameters, and we refer the reader to \citet{Schulman15TRPO} for more details.

{\bf Reward-Weighted Regression (RWR)} \cite{Peters07RWR, Kober09POWER}: This algorithm formulates the policy optimization as an Expectation-Maximization problem to avoid the need to manually choose learning rate, and the method is guaranteed to converge to a locally optimal solution. At each iteration, this algorithm optimizes a lower bound of the log-expected return: $\theta = \arg\max_{\theta'} \cL(\theta')$, where
$$\cL(\theta) = \frac{1}{NT} \sum_{i=1}^N \sum_{t=0}^T \log \pi(a_t^i|s_t^i; \theta) \rho(R_t^i - b_t^i)$$
Here, $\rho: \bR \rightarrow \bR_{\geq 0}$ is a function that transforms raw returns to nonnegative values. Following \citet{Deisenroth2013PSSurvey}, we choose $\rho$ to be $\rho(R) = R - R_{\text{min}}$, where $R_{\text{min}}$ is the minimum return among all trajectories collected in the current iteration.

{\bf Relative Entropy Policy Search (REPS)} \cite{Peters10REPS}: This algorithm limits the loss of information per iteration and aims to ensure a smooth learning progress \cite{Deisenroth2013PSSurvey}. At each iteration, we collect all trajectories into a dataset $\cD = \{(s_i, a_i, r_i, s_i') \}_{i=1}^M$, where $M$ is the total number of samples. Then, we first solve for the dual parameters $[\eta^*, \nu^*] = \arg\min_{\eta', \nu'} g(\eta', \nu')$ s.t. $\eta > 0$, where
$$g(\eta, \nu) = \eta\delta_{\text{KL}} + \eta \log \left( \frac{1}{M} \sum_{i=1}^M e^{\delta_i(\nu) / \eta}\right).$$
Here $\delta_{\text{KL}}> 0$ controls the step size of the policy, and $\delta_i(\nu) = r_i + \nu^T(\phi(s_i') - \phi(s_i))$ is the sample Bellman error. We then solve for the new policy parameters: $$\theta_{k+1} = \mathop{\arg\max}_{\theta} \frac{1}{M} \sum_{i=1}^M e^{\delta_i(\nu^*) / \eta^*} \log \pi(a_i | s_i; \theta).$$


{\bf Trust Region Policy Optimization (TRPO)} \cite{Schulman15TRPO}: 
This algorithm allows more precise control on the expected policy improvement than TNPG through the introduction of a surrogate loss.
At each iteration, we solve the following constrained optimization problem (replacing expectations with samples):
\begin{eqnarray*}
&\mathop{\textrm{maximize}}_{\theta}& \bE_{s \sim \rho_\thetaold, a \sim \pi_\thetaold} \left[ \frac{\pi_\theta(a|s)}{\pi_\thetaold(a|s)} A_\thetaold (s, a)\right]\cr
&\text{s.t.}& E_{s\sim \rho_{\thetaold}}[D_{\text{KL}}(\pi_\thetaold(\cdot |s) \| \pi_\theta(\cdot|s))] \leq \delta_{\text{KL}}
\end{eqnarray*}
where $\rho_\theta = \rho_{\pi_\theta}$ is the discounted state-visitation frequencies induced by $\pi_\theta$,
$A_{\theta_k}(s, a)$, known as the advantage function, is estimated by the empirical return minus the baseline, and $\delta_{\text{KL}}$ is a step size parameter which controls how much the policy is allowed to change per iteration. We follow the procedure described in the original paper for solving the optimization, which results in the same descent direction as TNPG with an extra line search in the objective and KL constraint.

{\bf Cross Entropy Method (CEM)} \cite{Rubinstein99CEM, Szita06CEM}: Unlike previously mentioned methods, which perform exploration through stochastic actions, CEM performs exploration directly in the policy parameter space. At each iteration, we produce $N$ perturbations of the policy parameter: $\theta_i \sim \cN(\mu_k, \Sigma_k)$, and perform a rollout for each sampled parameter. Then, we compute the new mean and diagonal covariance using the parameters that correspond to the top $q$-quantile returns.

{\bf Covariance Matrix Adaption Evolution Strategy (CMA-ES)} \cite{Hansen2001CMAES}: Similar to CEM, CMA-ES is a gradient-free evolutionary approach for optimizing nonconvex objective functions. In our case, this objective function equals the average sampled return. In contrast to CEM, CMA-ES estimates the covariance matrix of a multivariate normal distribution through incremental adaption along evolution paths, which contain information about the correlation between consecutive updates.

\subsection{Online Algorithms}

{\bf Deep Deterministic Policy Gradient (DDPG)}  \cite{Lillicrap15}: Compared to batch algorithms, the DDPG algorithm continuously improves the policy as it explores the environment. It applies gradient descent to the policy with minibatch data sampled from a replay pool, where the gradient is computed via
\[\widehat{\nabla_\theta \eta(\mu_\theta)} = \sum_{i=1}^B \left.\nabla_a Q_\phi(s_i, a)\right|_{a=\mu_\theta(s_i)} \nabla_\theta \mu_\theta(s_i)\]
where $B$ is the batch size. The critic $Q$ is trained via gradient descent on the $\ell^2$ loss of the Bellman error $L = \frac{1}{B}\sum_{i=1}^B (y_i - Q_\phi(s_i,a_i))^2$, where $y_i = r_i + \gamma Q'_{\phi'}(s'_i, \mu'_{\theta'}(s'_i))$. To improve stability of the algorithm, we use target networks for both the critic and the policy when forming the regression target $y_i$. We refer the reader to \citet{Lillicrap15} for a more detailed description of the algorithm.

\subsection{Recurrent Variants}
We implement direct applications of the aforementioned batch-based algorithms to recurrent policies. The only modification required is to replace $\pi(a_t^i|s_t^i)$ by $\pi(a_t^i|o_{1:t}^i, a_{1:t-1}^i)$, where $o_{1:t}^i$ and $a_{1:t-1}$ are the histories of past and current observations and past actions. Recurrent versions of reinforcement learning algorithms have been studied in many existing works, such as \citet{bakker2001reinforcement}, \citet{schafer2005solving}, \citet{wierstra2007solving}, and \citet{heess2015memory}.


\section{Experiment Setup}

In this section, we elaborate on the experimental setup used to generate the results.

{\bf Performance Metrics}: For each report unit (a particular algorithm running on a particular task), we define its performance as $\frac{1}{\sum_{i=1}^I N_i} \sum_{i=1}^I \sum_{n=1}^{N_i} R_{in}$, where $I$ is the number of training iterations, $N_i$ is the number of trajectories collected in the $i$th iteration, and $R_{in}$ is the undiscounted return for the $n$th trajectory of the $i$th iteration,

{\bf Hyperparameter Tuning}: For the DDPG algorithm, we used the hyperparametes reported in \citet{Lillicrap15}. For the other algorithms, we follow the approach in \cite{Mnih15}, and we select two tasks in each category, on which a grid search of hyperparameters is performed. Each choice of hyperparameters is executed under five random seeds. The criterion for the best hyperparameters is defined as $\mathrm{mean}(\mathrm{returns}) - \mathrm{std}(\mathrm{returns})$. This metric selects against large fluctuations of performance due to overly large step sizes. 

For the other tasks, we try both of the best hyperparameters found in the same category, and report the better performance of the two. This gives us insights into both the maximum possible performance when extensive hyperparameter tuning is performed, and the robustness of the best hyperparameters across different tasks.


\begin{sidewaystable*}[!p]
\footnotetext{$^a$Except for the hierarchical tasks}
\caption{Performance of the implemented algorithms in terms of average return over all training iterations for five different random seeds (same across all algorithms). The results of the best-performing algorithm on each task, as well as all algorithms that have performances that are not statistically significantly different (Welch's t-test with $p<0.05$), are highlighted in boldface.$^a$ In the tasks column, the partially observable variants of the tasks are annotated as follows: LS stands for limited sensors, NO for noisy observations and delayed actions, and SI for system identifications. The notation N/A denotes that an algorithm has failed on the task at hand, e.g., CMA-ES leading to out-of-memory errors in the Full Humanoid task.}
\label{table:main_eval}
\centering
\begin{scriptsize}
\setlength{\tabcolsep}{2pt}
\begin{tabular}{
        l
        S[table-format=5.1]@{\,\( \pm \)\,}
        S[table-format=1.1]
        S[table-format=5.1]@{\,\( \pm \)\,}
        S[table-format=3.1]
        S[table-format=5.1]@{\,\( \pm \)\,}
        S[table-format=3.1]
        S[table-format=5.1]@{\,\( \pm \)\,}
        S[table-format=3.1]
        S[table-format=5.1]@{\,\( \pm \)\,}
        S[table-format=3.1]
        S[table-format=5.1]@{\,\( \pm \)\,}
        S[table-format=3.1]
        S[table-format=5.1]@{\,\( \pm \)\,}
        S[table-format=3.1]
        S[table-format=5.1]@{\,\( \pm \)\,}
        S[table-format=3.1]
        S[table-format=5.1]@{\,\( \pm \)\,}
        S[table-format=3.1]
        }
        \sisetup{detect-weight=true,detect-inline-weight=math}

\\

\hline
\abovespace
Task & \multicolumn{2}{c}{Random} & \multicolumn{2}{c}{REINFORCE} & \multicolumn{2}{c}{TNPG} & \multicolumn{2}{c}{RWR} & \multicolumn{2}{c}{REPS} & \multicolumn{2}{c}{TRPO} & \multicolumn{2}{c}{CEM} & \multicolumn{2}{c}{CMA-ES} & \multicolumn{2}{c}{DDPG} \belowspace\\

\hline\\

Cart-Pole Balancing           & 77.1 & 0.0 & 4693.7 & 14.0 & {\bf 3986.4} & {\bf 748.9} & {\bf 4861.5} & {\bf 12.3}  & 565.6 & 137.6 & {\bf 4869.8} & {\bf 37.6} & 4815.4 & 4.8 & 2440.4 & 568.3 &  \ignore{ddpg} 4634.4 & 87.8 \\
Inverted Pendulum*           & -153.4 & 0.2 & 13.4 & 18.0 & {\bf 209.7} & {\bf 55.5} & 84.7 & 13.8 & -113.3 & 4.6 & {\bf 247.2} & {\bf 76.1} & 38.2 & 25.7 & -40.1 & 5.7 & \ignore{ddpg} 40.0 & 244.6 \\
Mountain Car           & -415.4 & 0.0 & -67.1 & 1.0 & {\bf -66.5} & {\bf 4.5} & -79.4 & 1.1 & -275.6 & 166.3 & {\bf -61.7} & {\bf 0.9}  & -66.0 & 2.4 &  -85.0 & 7.7 & -288.4 & 170.3  \\
Acrobot           & -1904.5 & 1.0 & -508.1 & 91.0 & -395.8 &  121.2 &  -352.7 &  35.9 & -1001.5 & 10.8 &  -326.0 &  24.4  & -436.8 & 14.7 & -785.6 & 13.1 & {\bf -223.6} & {\bf 5.8}  \\
Double Inverted Pendulum*           & 149.7 & 0.1 & 4116.5 & 65.2 & {\bf 4455.4} & {\bf 37.6} & 3614.8 & 368.1 & 446.7 & 114.8 & {\bf 4412.4} & {\bf 50.4}  & 2566.2 & 178.9 & 1576.1 & 51.3 & 2863.4 & 154.0
\belowspace\\

\hline
\abovespace

Swimmer*           & -1.7 & 0.1 & 92.3 & 0.1 & {\bf 96.0} & {\bf 0.2} & 60.7 & 5.5 & 3.8 & 3.3 & {\bf 96.0} & {\bf 0.2}  & 68.8 & 2.4 & 64.9 & 1.4 & 85.8 & 1.8  \\
Hopper           & 8.4 & 0.0 & 714.0 & 29.3 & {\bf 1155.1} & {\bf 57.9} & 553.2 & 71.0 & 86.7 & 17.6 & {\bf 1183.3} & {\bf 150.0} & 63.1 & 7.8 & 20.3 & 14.3 & 267.1 & 43.5  \\
2D Walker           & -1.7 & 0.0 & 506.5 & 78.8 & {\bf 1382.6} & {\bf 108.2}  & 136.0 & 15.9 & -37.0 & 38.1 & {\bf 1353.8} & {\bf 85.0} & 84.5 & 19.2 & 77.1 & 24.3 & 318.4 & 181.6 \\
Half-Cheetah           & -90.8 & 0.3 & 1183.1 & 69.2 & {\bf 1729.5} & {\bf 184.6} & 376.1 & 28.2 & 34.5 & 38.0 & {\bf 1914.0} & {\bf 120.1}  & 330.4 & 274.8 & 441.3 & 107.6 & {\bf 2148.6} & {\bf 702.7} \\
Ant*           & 13.4 & 0.7 & 548.3 & 55.5 & {\bf 706.0} & {\bf 127.7}  & 37.6 & 3.1 & 39.0 & 9.8 & {\bf 730.2} & {\bf 61.3} & 49.2 & 5.9 & 17.8 & 15.5 & 326.2 & 20.8 \\
Simple Humanoid           & 41.5 & 0.2 & 128.1 & 34.0 & {\bf 255.0} & {\bf 24.5} & 93.3 & 17.4 & 28.3 & 4.7 & {\bf 269.7} & {\bf 40.3} & 60.6 & 12.9 & 28.7 & 3.9 & 99.4 & 28.1 \\
Full Humanoid           & 13.2 & 0.1 & 262.2 & 10.5 & {\bf 288.4} & {\bf 25.2} & 46.7 & 5.6 & 41.7 & 6.1 & {\bf 287.0} & {\bf 23.4} & 36.9 & 2.9 & \text{\textnormal{N/A}\hspace{-15pt}} & \text{N/A}  &  119.0 & 31.2  \belowspace\\

\hline
\abovespace

Cart-Pole Balancing (LS)*           & 77.1 &0.0 & 420.9 & 265.5 & {\bf 945.1} & {\bf 27.8} & 68.9 & 1.5 & 898.1 & 22.1 & {\bf 960.2} & {\bf 46.0}  & 227.0 & 223.0 & 68.0 & 1.6 \\
Inverted Pendulum (LS)           & -122.1 &0.1 & -13.4 & 3.2 & {\bf 0.7} & {\bf 6.1} & -107.4 & 0.2 & -87.2 & 8.0 & {\bf 4.5} & {\bf 4.1}  & -81.2 & 33.2 & -62.4 & 3.4  \\
Mountain Car (LS)           & -83.0 &0.0 & -81.2 & 0.6 & {\bf -65.7} &\bfseries  9.0 & -81.7 & 0.1 & -82.6 & 0.4 & {\bf -64.2} & {\bf 9.5}  & {\bf -68.9} & {\bf 1.3} & {\bf -73.2} & {\bf 0.6}  \\
Acrobot (LS)*           & -393.2 &0.0 & -128.9 & 11.6 & {\bf -84.6} & {\bf 2.9} & -235.9 & 5.3 & -379.5 & 1.4 & {\bf -83.3} & {\bf 9.9}  & -149.5 & 15.3 & -159.9 & 7.5  \belowspace\\

\hline
\abovespace

Cart-Pole Balancing (NO)*           & 101.4 &0.1 & 616.0 & 210.8 & {\bf 916.3} & {\bf 23.0}  & 93.8 & 1.2 & 99.6 & 7.2 & 606.2 & 122.2 & 181.4 & 32.1 & 104.4 & 16.0  \\
Inverted Pendulum (NO)           & -122.2 &0.1 & 6.5 & 1.1 & {\bf 11.5} & {\bf 0.5}  & -110.0 & 1.4 & -119.3 & 4.2 & {\bf 10.4} & {\bf 2.2} & -55.6 & 16.7 & -80.3 & 2.8  \\
Mountain Car (NO)           & -83.0 &0.0 & -74.7 & 7.8 & {\bf -64.5} & {\bf 8.6} & -81.7 & 0.1 & -82.9 & 0.1 & {\bf -60.2} & {\bf 2.0}  & -67.4 & 1.4 & -73.5 & 0.5  \\
Acrobot (NO)*           & -393.5 &0.0 & {\bf -186.7} & {\bf 31.3} & {\bf -164.5} & {\bf 13.4} & -233.1 & 0.4 & -258.5 & 14.0 & {\bf -149.6} & {\bf 8.6}  & -213.4 & 6.3 & -236.6 & 6.2   \belowspace\\

\hline
\abovespace

Cart-Pole Balancing (SI)*  & 76.3 & 0.1 & 431.7 & 274.1 & {\bf 980.5} & {\bf 7.3}  & 69.0 & 2.8 & 702.4 & 196.4 & {\bf 980.3} & {\bf 5.1} & 746.6 & 93.2 & 71.6 & 2.9  \\
Inverted Pendulum (SI)           & -121.8 &0.2 & -5.3 & 5.6 & {\bf 14.8} & {\bf 1.7}  & -108.7 & 4.7 & -92.8 & 23.9 & {\bf 14.1} & {\bf 0.9} & -51.8 & 10.6 & -63.1 & 4.8  \\
Mountain Car (SI)           & -82.7 &0.0 & -63.9 & 0.2 & {\bf -61.8} & {\bf 0.4} & -81.4 & 0.1 & -80.7 & 2.3 & {\bf -61.6} & {\bf 0.4}  & -63.9 & 1.0 & -66.9 & 0.6  \\
Acrobot (SI)*       & -387.8 & 1.0    & {\bf -169.1} & {\bf 32.3} & {\bf -156.6} & {\bf 38.9}  & -233.2 & 2.6 & -216.1 & 7.7 & {\bf -170.9} & {\bf 40.3} & -250.2 & 13.7 & -245.0 & 5.5   \belowspace\\
\hline
\abovespace

Swimmer + Gathering           & 0.0 & 0.0 & 0.0 & 0.0 & 0.0 & 0.0 & 0.0 & 0.0 & 0.0 & 0.0 & 0.0 & 0.0 &  0.0 &  0.0  & 0.0 & 0.0 & 0.0 & 0.0 \\ 
Ant + Gathering           & -5.8 & 5.0 &  -0.1 &  0.1  & -0.4 & 0.1 & -5.5 & 0.5 & -6.7 & 0.7 & -0.4 & 0.0 & -4.7 & 0.7 & \text{\textnormal{N/A}\hspace{-15pt}} & \text{N/A} &-0.3 & 0.3 \\ 
Swimmer + Maze           & 0.0 & 0.0 &  0.0 &  0.0  &  0.0 &  0.0  &  0.0 &  0.0  &  0.0 &  0.0  &  0.0 &  0.0  &  0.0 &  0.0  &  0.0 &  0.0 &  0.0 &  0.0  \\ 
Ant + Maze           & 0.0 & 0.0 &  0.0 &  0.0  &  0.0 &  0.0  &  0.0 &  0.0  &  0.0 &  0.0  &  0.0 &  0.0  &  0.0 &  0.0  & \text{\textnormal{N/A}\hspace{-15pt}} & \text{N/A}  &  0.0 &  0.0 \belowspace\\

\hline
\end{tabular}
\end{scriptsize}
\end{sidewaystable*}

{\bf Policy Representation}: For basic, locomotion, and hierarchical tasks and for batch algorithms, we use a feed-forward neural network policy with 3 hidden layers, consisting of $100$, $50$, and $25$ hidden units with tanh nonlinearity at the first two hidden layers, which map each state to the mean of a Gaussian distribution. The log-standard deviation is parameterized by a global vector independent of the state, as done in \citet{Schulman15TRPO}. For all partially observable tasks, we use a recurrent neural network with a single hidden layer consisting of $32$ LSTM hidden units \cite{hochreiter1997long}.

For the DDPG algorithm which trains a deterministic policy, we follow \citet{Lillicrap15}. For both the policy and the $Q$ function, we use the same architecture of a feed-forward neural network with 2 hidden layers, consisting of $400$ and $300$ hidden units with relu activations.


{\bf Baseline}: For all gradient-based algorithms except REPS, we can subtract a baseline from the empirical return to reduce variance of the optimization. We use a linear function as the baseline with a time-varying feature vector.

\label{section:experiment-setup}

\begin{figure*}[!t]
\subfigure[]{
\includegraphics[width=117px]{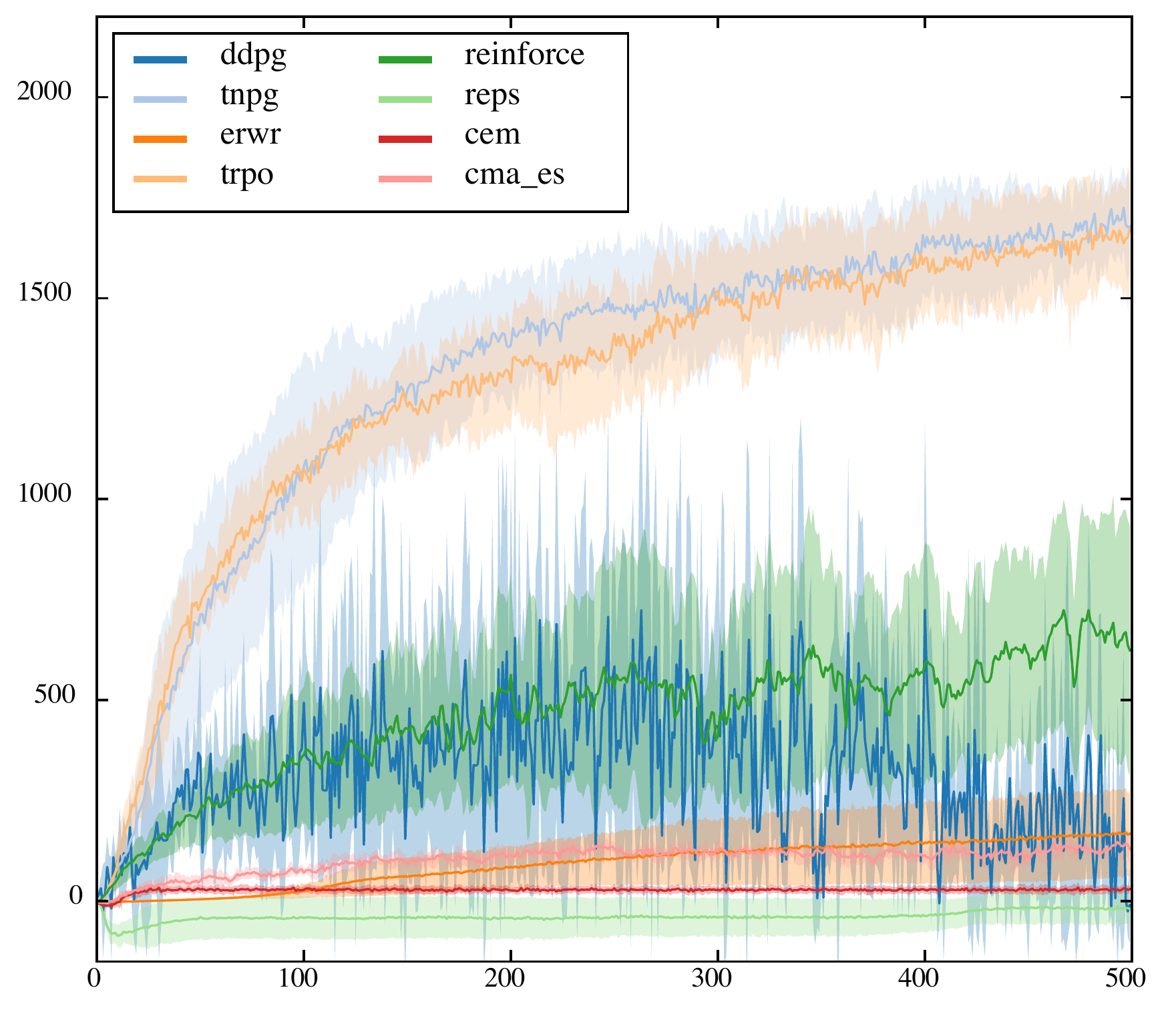}\label{fig:walker_curve}}
\subfigure[]{
\includegraphics[width=117px]{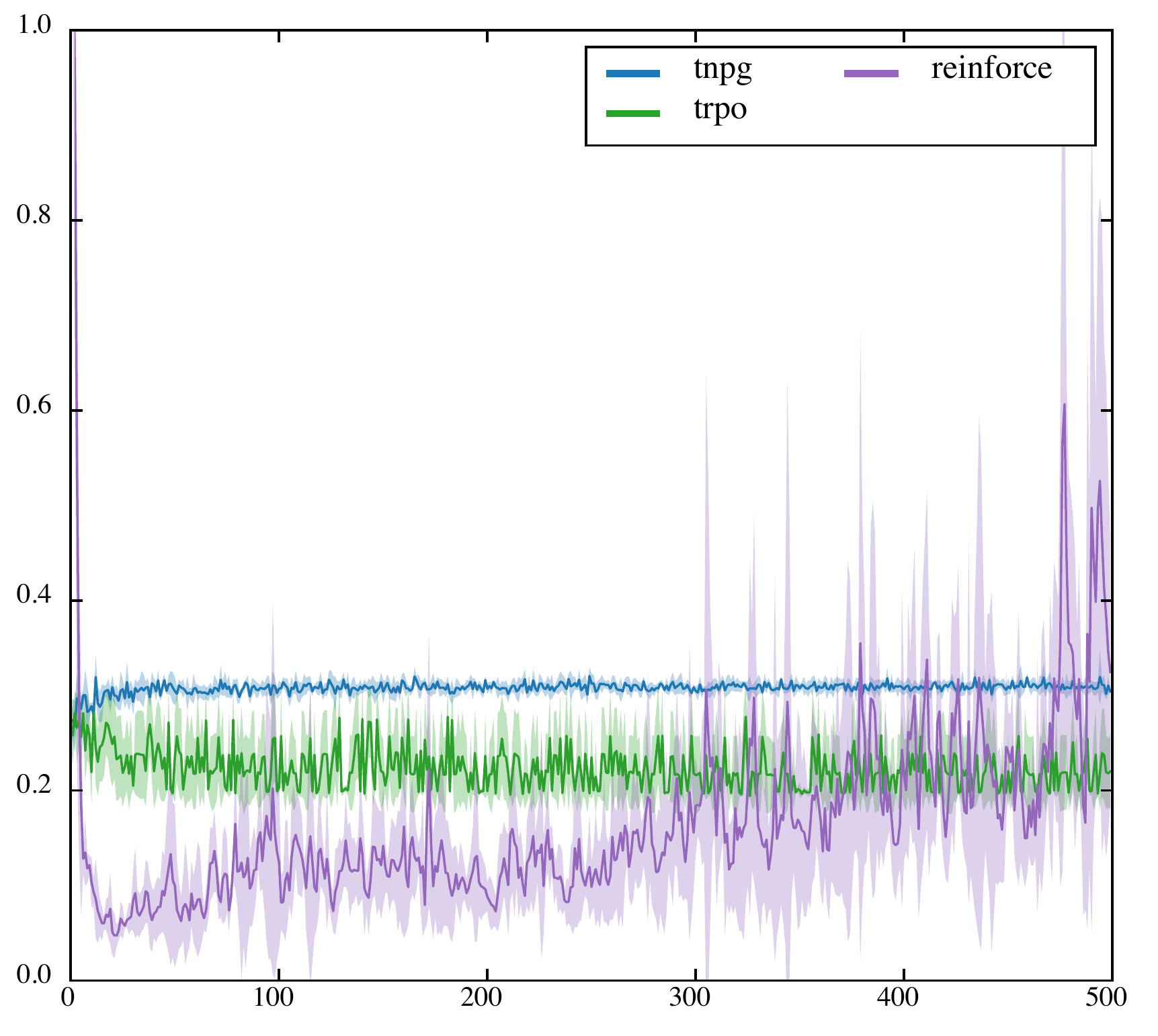}\label{fig:walker_kl}}
\subfigure[]{   
\includegraphics[width=117px]{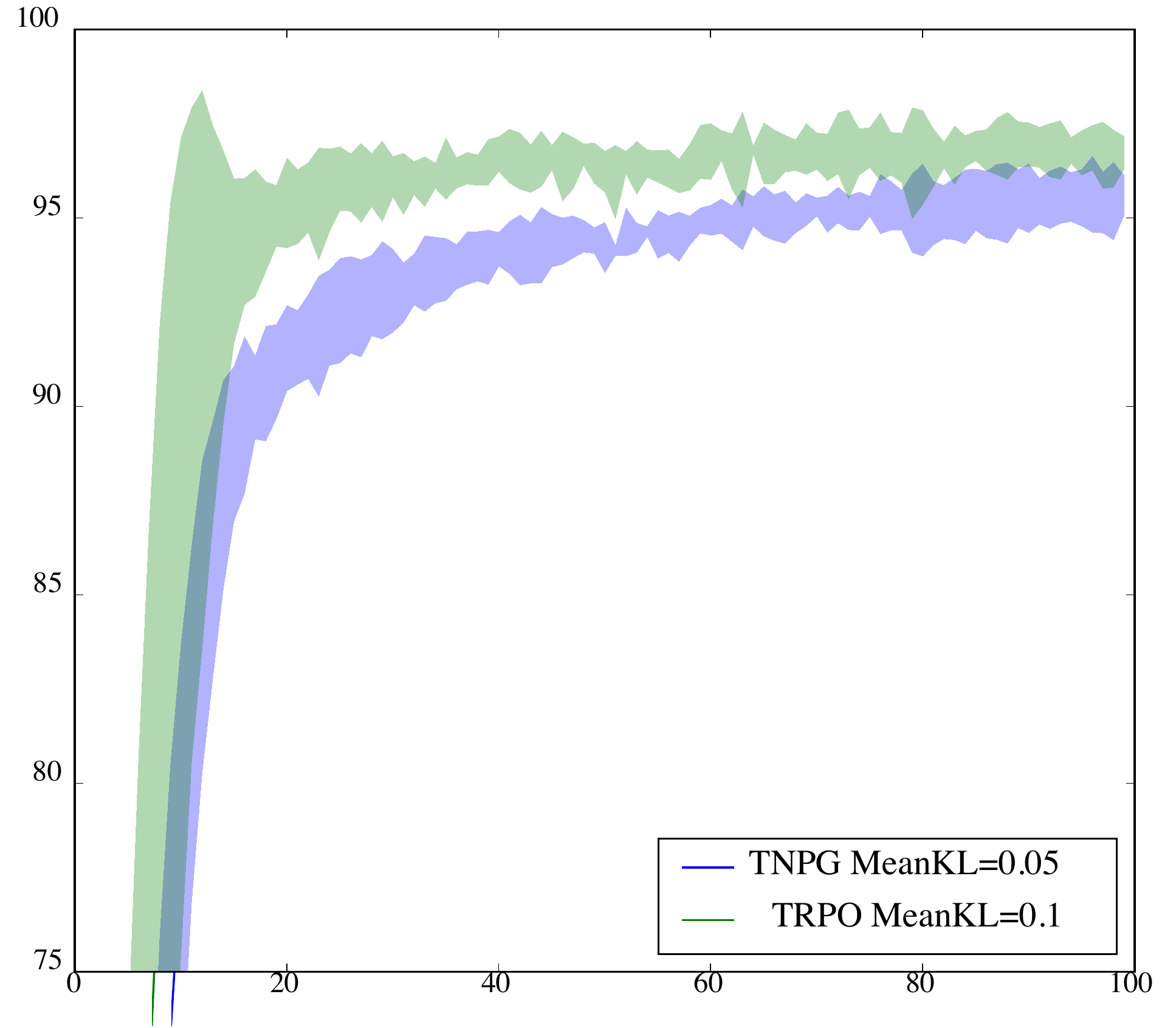}\label{fig:swimmer_npg_trpo}}
\subfigure[]{
\includegraphics[width=117px]{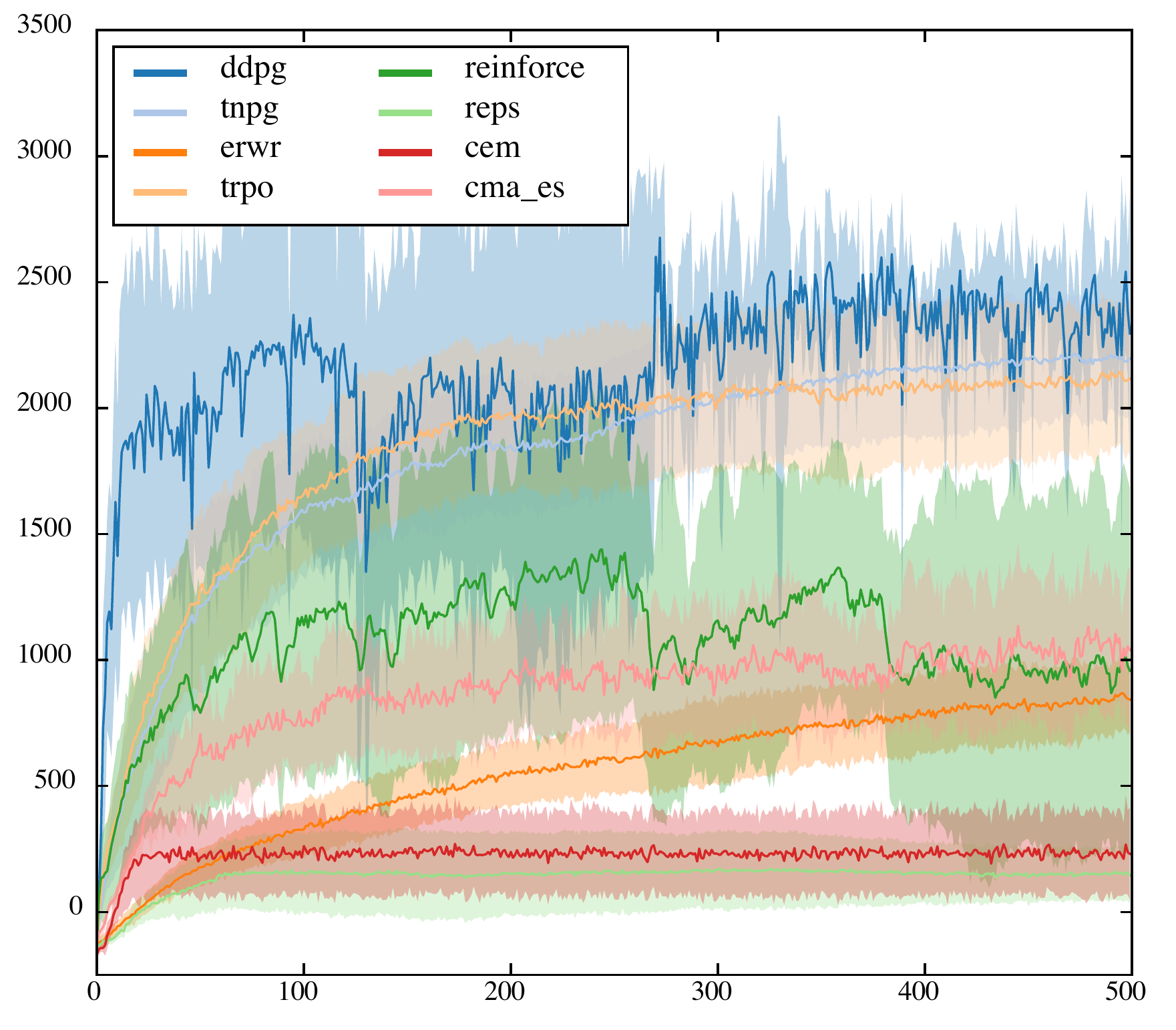}\label{fig:cheetah_curve}}
\caption{Performance as a function of the number of iterations; the shaded area depicts the mean $\pm$ the standard deviation over five different random seeds: \subref{fig:walker_curve} Performance comparison of all algorithms in terms of the average reward on the Walker task; \subref{fig:walker_kl} Comparison between REINFORCE, TNPG, and TRPO in terms of the mean KL-divergence on the Walker task; \subref{fig:swimmer_npg_trpo} Performance comparison on TNPG and TRPO on the Swimmer task; \subref{fig:cheetah_curve} Performance comparison of all algorithms in terms of the average reward on the Half-Cheetah task.}\label{fig:test}
\end{figure*}

\section{Results and Discussion}

\label{section:results}

The main evaluation results are presented in Table~\ref{table:main_eval}. The tasks on which the grid search is performed are marked with (*). In each entry, the pair of numbers shows the mean and standard deviation of the normalized cumulative return using the best possible hyperparameters.

{\bf REINFORCE:} Despite its simplicity, REINFORCE is an effective algorithm in optimizing deep neural network policies in most basic and locomotion tasks. Even for high-DOF tasks like Ant, REINFORCE can achieve competitive results. However we observe that REINFORCE sometimes suffers from premature convergence to local optima as noted by \citet{peters2008reinforcement}, which explains the performance gaps between REINFORCE and TNPG on tasks such as Walker (Figure~\ref{fig:walker_curve}). By visualizing the final policies, we can see that REINFORCE results in policies that tend to jump forward and fall over to maximize short-term return instead of acquiring a stable walking gait to maximize long-term return. In Figure~\ref{fig:walker_kl}, we can observe that even with a small learning rate, steps taken by REINFORCE can sometimes result in large changes to policy distribution, which may explain the fast convergence to local optima. 

{\bf TNPG and TRPO:} Both TNPG and TRPO outperform other batch algorithms by a large margin on most tasks, confirming that constraining the change in the policy distribution results in more stable learning \cite{peters2008reinforcement}.


Compared to TNPG, TRPO offers better control over each policy update by performing a line search in the natural gradient direction to ensure an improvement in the surrogate loss function. 
We observe that hyperparameter grid search tends to select conservative step sizes ($\delta_{\text{KL}}$) for TNPG, which alleviates the issue of performance collapse caused by a large update to the policy. By contrast, TRPO can robustly enforce constraints with larger a $\delta_{\text{KL}}$ value and hence speeds up learning in some cases. For instance, grid search on the Swimmer task reveals that the best step size for TNPG is $\delta_{\text{KL}} = 0.05$, whereas TRPO's best step-size is larger: $\delta_{\text{KL}} = 0.1$. As shown in Figure~\ref{fig:swimmer_npg_trpo}, this larger step size enables slightly faster learning.



{\bf RWR:} RWR is the only gradient-based algorithm we implemented that does not require any hyperparameter tuning. It can solve some basic tasks to a satisfactory degree, but fails to solve more challenging tasks such as locomotion. We observe empirically that RWR shows fast initial improvement followed by significant slow-down, as shown in Figure~\ref{fig:cheetah_curve}.


{\bf REPS:} Our main observation is that REPS is especially prone to early convergence to local optima in case of continuous states and actions. Its final outcome is greatly affected by the performance of the initial policy, an observation that is consistent with the original work of \citet{Peters10REPS}. This leads to a bad performance on average, although under particular initial settings the algorithm can perform on par with others. Moreover, the tasks presented here do not assume the existence of a stationary distribution, which is assumed in \citet{Peters10REPS}. In particular, for many of our tasks, transient behavior is of much greater interest than steady-state behavior, which agrees with previous observation by \citet{HofNeuPet15}, 

{\bf Gradient-free methods:} Surprisingly, even when training deep neural network policies with thousands of parameters, CEM achieves very good performance on certain basic tasks such as Cart-Pole Balancing and Mountain Car, suggesting that the dimension of the searching parameter is not always the limiting factor of the method. However, the performance degrades quickly as the system dynamics becomes more complicated. We also observe that CEM outperforms CMA-ES, which is remarkable as CMA-ES estimates the full covariance matrix. For higher-dimensional policy parameterizations, the computational complexity and memory requirement for CMA-ES become noticeable. On tasks with high-dimensional observations, such as the Full Humanoid, the CMA-ES algorithm runs out of memory and fails to yield any results, denoted as N/A in Table~\ref{table:main_eval}.

{\bf DDPG:} Compared to batch algorithms, we found that DDPG was able to converge significantly faster on certain tasks like Half-Cheetah due to its greater sample efficiency. However, it was less stable than batch algorithms, and the performance of the policy can degrade significantly during training. We also found it to be more susceptible to scaling of the reward. In our experiment for DDPG, we rescaled the reward of all tasks by a factor of $0.1$, which seems to improve the stability.


{\bf Partially Observable Tasks:} We experimentally verify that recurrent policies can find better solutions than feed-forward policies in Partially Observable Tasks but recurrent policies are also more difficult to train. As shown in Table~\ref{table:main_eval}, derivative-free algorithms like CEM and CMA-ES work considerably worse with recurrent policies. Also we note that the performance gap between REINFORCE and TNPG widens when they are applied to optimize recurrent policies, which can be explained by the fact that a small change in parameter space can result in a bigger change in policy distribution with recurrent policies than with feedforward policies.

{\bf Hierarchical Tasks:} We observe that all of our implemented algorithms achieve poor performance on the hierarchical tasks, even with extensive hyperparameter search and $500$ iterations of training. It is an interesting direction to develop algorithms that can automatically discover and exploit the hierarchical structure in these tasks.



\section{Related Work}
\label{section:related_work}

In this section, we review existing benchmarks of continuous control tasks. The earliest efforts of evaluating reinforcement learning algorithms started in the form of individual control problems described in symbolic form. Some widely adopted tasks include the inverted pendulum \cite{stephenson1908xx, donaldson1960error, widrow1964pattern}, mountain car \cite{Moore90MountainCar}, and Acrobot \cite{dejong1994swinging}.
These problems are frequently incorporated into more comprehensive benchmarks.

Some reinforcement learning benchmarks contain low-dimensional continuous control tasks, such as the ones introduced above, including RLLib \cite{RLLibCapital}, MMLF \cite{MMLF}, RL-Toolbox \cite{RLToolbox},
JRLF \cite{JRLF},
Beliefbox \cite{BeliefBox}, Policy Gradient Toolbox \cite{PolicyGradientToolbox}, and ApproxRL \cite{ApproxRL}. A series of RL competitions has also been held in recent years \cite{dutech2005reinforcement, dimitrakakis2014reinforcement}, again with relatively low-dimensional actions. In contrast, our benchmark contains a wider range of tasks with high-dimensional continuous state and action spaces.

Previously, other benchmarks have been proposed for high-dimensional control tasks. Tdlearn \cite{tdlearn} includes a 20-link pole balancing task, DotRL \cite{dotRL} includes a variable-DOF octopus arm and a 6-DOF planar cheetah model, PyBrain \cite{PyBrain} includes a 16-DOF humanoid robot with standing and jumping tasks, RoboCup Keepaway \cite{stone2005keepaway} is a multi-agent game which can have a flexible dimension of actions by varying the number of agents, and SkyAI \cite{SkyAI} includes a 17-DOF humanoid robot with crawling and turning tasks. Other libraries such as CL-Square \cite{CLSquare} and RLPark \cite{RLPark}
provide interfaces to actual hardware, e.g., Bioloid and iRobot Create. In contrast to these aforementioned testbeds, our benchmark makes use of simulated environments to reduce computation time and to encourage experimental reproducibility. Furthermore, it provides a much larger collection of tasks of varying difficulty.


\section{Conclusion}
\label{section:conclusion}

In this work, a benchmark of continuous control problems for reinforcement learning is presented, covering a wide variety of challenging tasks. We implemented several reinforcement learning algorithms, and presented them in the context of general policy parameterizations. Results show that among the implemented algorithms, TNPG, TRPO, and DDPG are effective methods for training deep neural network policies. Still, the poor performance on the proposed hierarchical tasks calls for new algorithms to be developed. Implementing and evaluating existing and newly proposed algorithms will be our continued effort. By providing an open-source release of the benchmark, we encourage other researchers to evaluate their algorithms on the proposed tasks.

\section*{Acknowledgements}

We thank Emo Todorov and Yuval Tassa for providing the MuJoCo simulator, and Sergey Levine, Aviv Tamar, Chelsea Finn, and the anonymous ICML reviewers for insightful comments. We also thank Shixiang Gu and Timothy Lillicrap for helping us diagnose the DDPG implementation. This work was supported in part by DARPA, the Berkeley Vision and
Learning Center (BVLC), the Berkeley Artificial Intelligence Research
(BAIR) laboratory, and Berkeley Deep Drive (BDD). Rein Houthooft is supported by a Ph.D. Fellowship of the Research Foundation - Flanders (FWO).




\setlength{\bibsep}{4.0pt}
\small
\bibliography{benchmark}
\bibliographystyle{icml2016}

\normalsize
\include{supplementary_text}

\end{document}

%% file: defs.tex
\usepackage{xifthen}
\usepackage{amsfonts}
\usepackage{amsopn}

\newcommand{\todo}[1][]{
\ifthenelse{\equal{#1}{}}{TODO}{TODO: #1}
}

\newcommand{\cS}{\mathcal{S}}
\newcommand{\cO}{\mathcal{O}}
\newcommand{\cA}{\mathcal{A}}
\newcommand{\cD}{\mathcal{D}}
\newcommand{\cN}{\mathcal{N}}
\newcommand{\cL}{\mathcal{L}}
\newcommand{\bR}{\mathbb{R}}
\newcommand{\bE}{\mathbb{E}}
\newcommand{\thetaold}{{\theta_k}}

\newcommand{\norm}[1]{\|#1\|}

%% file: supplementary_text.tex
\renewcommand*{\UrlFont}{\normalsize}

\onecolumn

\icmltitle{Supplementary Material}
\setcounter{section}{0}

\section{Task Specifications}

Below we provide some specifications for the task observations, actions, and rewards. Please refer to the benchmark source code (\url{https://github.com/rllab/rllab}) for complete specification of physics parameters.

\subsection{Basic Tasks}

{\bf Cart-Pole Balancing}: In this task, an inverted pendulum is mounted on a pivot point on a cart. The cart itself is restricted to linear movement, achieved by applying horizontal forces. Due to the system's inherent instability, continuous cart movement is needed to keep the pendulum upright. The observation consists of the cart position $x$, pole angle $\theta$, the cart velocity $\dot x$, and the pole velocity $\dot\theta$. The 1D action consists of the horizontal force applied to the cart body. The reward function is given by 
$r(s, a) := 10 - (1 - \cos(\theta)) - 10^{-5} \norm{a}_2^2$. The episode terminates when $|x| > 2.4$ or $|\theta| > 0.2$.

{\bf Cart-Pole Swing Up}: This is a more complicated version of the previous task, in which the system should not only be able to balance the pole, but first succeed in swinging it up into an upright position. This task extends the working range of the inverted pendulum to \SI{360}{\degree}. This is a nonlinear extension of the previous task. It has the same observation and action as in balancing. The reward function is given by $r(s, a) := \cos(\theta)$. The episode terminates when $|x| > 3$, with a penalty of $-100$.

{\bf Mountain Car}: In this task, a car has to escape a valley by repetitive application of tangential forces. Because the maximal tangential force is limited, the car has to alternately drive up along the two slopes of the valley in order to build up enough inertia to overcome gravity. This brings a challenge of exploration, since before first reaching the goal among all trials, a locally optimal solution exists, which is to drive to the point closest to the target and stay there for the rest of the episode. The observation is given by the horizontal position $x$ and the horizontal velocity $\dot x$ of the car. The reward is given by $r(s, a) := -1 + \textrm{height}$, with $\textrm{height}$ the car's vertical offset. The episode terminates when the car reaches a target height of $0.6$. Hence the goal is to reach the target as soon as possible.

{\bf Acrobot Swing Up}: In this task, an under-actuated, two-link robot has to swing itself into an upright position. It consists of two joints of which the first one has a fixed position and only the second one can exert torque. The goal is to swing the robot into an upright position and stabilize around that position. The controller not only has to swing the pendulum in order to build up inertia, similar to the Mountain Car task, but also has to decelerate it in order to prevent it from tipping over. The observation includes the two joint angles, $\theta_1$ and $\theta_2$, and their velocities, $\dot\theta_1$ and $\dot\theta_2$. The action is the torque applied at the second joint. The reward is defined as $r(s, a) := -\norm{\mathrm{tip}(s) - \mathrm{tip}_{\mathrm{target}}}_2$, where $\mathrm{tip}(s)$ computes the Cartesian position of the tip of the robot given the joint angles. No termination condition is applied.

{\bf Double Inverted Pendulum Balancing}: This task extends the Cart-Pole Balancing task by replacing the single-link pole by a two-link rigid structure. As in the former task, the goal is to stabilize the two-link pole near the upright position. This task is more difficult than single-pole balancing, since the system is even more unstable and requires the controller to actively maintain balance. The observation includes the cart position $x$, joint angles ($\theta_1$ and $\theta_2$), and joint velocities ($\dot\theta_1$ and $\dot\theta_2$). We encode each joint angle as its sine and cosine values. The action is the same as in cart-pole tasks. The reward is given by $r(s, a) = 10 - 0.01 x_{\mathrm{tip}}^2 - (y_{\mathrm{tip}} - 2)^2 - 10^{-3} \cdot \dot\theta_1^2 - 5\cdot 10^{-3} \cdot \dot\theta_2^2$, where $x_{\mathrm{tip}}, y_{\mathrm{tip}}$ are the coordinates of the tip of the pole.
No termination condition is applied. The episode is terminated when $y_{\text{tip}} \leq 1$.

\subsection{Locomotion Tasks}

{\bf Swimmer}: The swimmer is a planar robot with 3 links and 2 actuated joints. Fluid is simulated through viscosity forces, which apply drag on each link, allowing the swimmer to move forward. This task is the simplest of all locomotion tasks, since there are no irrecoverable states in which the swimmer can get stuck, unlike other robots which may fall down or flip over. This places less burden on exploration. The $13$-dim observation includes the joint angles, joint velocities, as well as the coordinates of the center of mass. The reward is given by $r(s, a) = v_x - 0.005 \norm{a}_2^2$, where $v_x$ is the forward velocity. No termination condition is applied.

{\bf Hopper}: The hopper is a planar monopod robot with 4 rigid links, corresponding to the torso, upper leg, lower leg, and foot, along with 3 actuated joints. More exploration is needed than the swimmer task, since a stable hopping gait has to be learned without falling. Otherwise, it may get stuck in a local optimum of diving forward. The $20$-dim observation includes joint angles, joint velocities, the coordinates of center of mass, and constraint forces. The reward is given by
$r(s, a) := v_x - 0.005 \cdot \norm{a}_2^2 + 1$,
where the last term is a bonus for being ``alive.'' The episode is terminated when $z_{body} < 0.7$ where $z_{body}$ is the $z$-coordinate of the body, or when $|\theta_y| < 0.2$, where $\theta_y$ is the forward pitch of the body.

{\bf Walker}: The walker is a planar biped robot consisting of 7 links, corresponding to two legs and a torso, along with 6 actuated joints. This task is more challenging than hopper, since it has more degrees of freedom, and is also prone to falling. The $21$-dim observation includes joint angles, joint velocities, and the coordinates of center of mass. The reward is given by
$r(s, a) := v_x - 0.005 \cdot \norm{a}_2^2$.
The episode is terminated when $z_{body} < 0.8$, $z_{body} > 2.0$, or when $|\theta_y| > 1.0$.

{\bf Half-Cheetah}: The half-cheetah is a planar biped robot with  9 rigid links, including two legs and a torso, along with 6 actuated joints. The $20$-dim observation includes joint angles, joint velocities, and the coordinates of the center of mass. The reward is given by
$r(s, a) = v_x - 0.05 \cdot \norm{a}_2^2$.
No termination condition is applied.

{\bf Ant}: The ant is a quadruped with 13 rigid links, including four legs and a torso, along with 8 actuated joints. This task is more challenging than the previous tasks due to the higher degrees of freedom. The $125$-dim observation includes joint angles, joint velocities, coordinates of the center of mass, a (usually sparse) vector of contact forces, as well as the rotation matrix for the body. The reward is given by $r(s, a) = v_x - 0.005 \cdot \norm{a}_2^2 - C_{\mathrm{contact}} + 0.05$, where $C_{\mathrm{contact}}$ penalizes contacts to the ground, and is given by
$5\cdot 10^{-4} \cdot \norm{F_{\mathrm{contact}}}_2^2$, where $F_{\mathrm{contact}}$ is the contact force vector clipped to values between $-1$ and $1$. The episode is terminated when $z_{body} < 0.2$ or when $z_{body} > 1.0$.

{\bf Simple Humanoid}: This is a simplified humanoid model with 13 rigid links, including the head, body, arms, and legs, along with 10 actuated joints. The increased difficulty comes from the increased degrees of freedom as well as the need to maintain balance. The $102$-dim observation includes the joint angles, joint velocities, vector of contact forces, and the coordinates of the center of mass. The reward is given by
$r(s, a) = v_x - 5\cdot 10^{-4} \norm{a}_2^2 - C_{\mathrm{contact}} - C_{\mathrm{deviation}} + 0.2$, where $C_{\mathrm{contact}} = 5\cdot 10^{-6} \cdot \norm{F_{\mathrm{contact}}}$, and $C_{\mathrm{deviation}} = 5\cdot 10^{-3} \cdot (v_y^2 + v_z^2)$ to penalize deviation from the forward direction. The episode is terminated when $z_{body} < 0.8$ or when $z_{body} > 2.0$.

{\bf Full Humanoid}: This is a humanoid model with 19 rigid links and 28 actuated joints. It has more degrees of freedom below the knees and elbows, which makes the system higher-dimensional and harder for learning. The $142$-dim observation includes the joint angles, joint velocities, vector of contact forces, and the coordinates of the center of mass. The reward and termination condition is the same as in the Simple Humanoid model.

\subsection{Partially Observable Tasks}

{\bf Limited Sensors}: The full description is included in the main text.

{\bf Noisy Observations and Delayed Actions}: For all tasks, we use a Gaussan noise with $\sigma=0.1$. The time delay is as follows: Cart-Pole Balancing 0.15 sec, Cart-Pole Swing Up 0.15 sec, Mountain Car 0.15 sec, Acrobot Swing Up 0.06 sec, and Double Inverted Pendulum Balancing 0.06 sec. This corresponds to $3$ discretization frames for each task.

{\bf System Identifications}: For Cart-Pole Balancing and Cart-Pole Swing Up, the pole length is varied uniformly between, 50\% and 150\%. For Mountain Car, the width of the valley varies uniformly between 75\% and 125\%. For Acrobot Swing Up, each of the pole length varies uniformly between 50\% and 150\%. For Double Inverted Pendulum Balancing, each of the pole length varies uniformly between 83\% and 167\%. Please refer to the benchmark source code for reference values.

\subsection{Hierarchical Tasks}

{\bf Locomotion + Food Collection}: During each episode, $8$ food units and $8$ bombs are placed in the environment. Collecting a food unit gives $+1$ reward, and collecting a bomb gives $-1$ reward. Hence the best cumulative reward for a given episode is $8$.

{\bf Locomotion + Maze}: During each episode, a $+1$ reward is given when the robot reaches the goal. Otherwise, the robot receives a zero reward throughout the episode. 

\section{Experiment Parameters}

For all batch gradient-based algorithms, we use the same time-varying feature encoding for the linear baseline:
\[ \phi_{s,t} = \mathrm{concat}(s, s \odot s, 0.01t, (0.01t)^2, (0.01t)^3, 1) \]
where $s$ is the state vector and $\odot$ represents element-wise product.

Table~\ref{experiment_setup} shows the experiment parameters for all four categories. We will then detail the hyperparameter search range for the selected tasks and report best hyperparameters, shown in Tables \ref{hyper_vpg}, \ref{hyper_npg}, \ref{hyper_trpo}, \ref{hyper_reps}, \ref{hyper_cem}, and \ref{hyper_cma_es}.

\begin{table}[!h]
\centering
\caption{Experiment Setup}
\label{experiment_setup}
\begin{tabular}{l|llll}
\cline{1-4} \\ [-8pt]
                     & Basic \& Locomotion       & Partially Observable & Hierarchical &  \\ [2pt] \cline{1-4} \\ [-8pt]
Sim. steps per Iter. & 50,000                    & 50,000               & 50,000       &  \\
Discount($\lambda$)  & 0.99                      & 0.99                 & 0.99         &  \\
Horizon              & 500                       & 100                  & 500          &  \\
Num. Iter.           & 500  & 300                  & 500        & \\ [2pt] \cline{1-4}
\end{tabular}
\end{table}

\begin{table}[!h]
\centering
\caption{Learning Rate $\alpha$ for REINFORCE}
\label{hyper_vpg}
\begin{tabular}{l|lll}
\cline{1-4} \\ [-8pt]
& Search Range              & Best                 &  \\ [2pt] \cline{1-4} \\ [-8pt]
Cart-Pole Swing Up     & $[1\times 10^{-4}, 1\times 10^{-1}]$            & $5\times 10^{-3}$               &  \\
Double Inverted Pendulum     & $[1\times 10^{-4}, 1\times 10^{-1}]$     & $5\times 10^{-3}$               &  \\
Swimmer               & $[1\times 10^{-4}, 1\times 10^{-1}]$            & $1\times 10^{-2}$               &  \\
Ant                   & $[1\times 10^{-4}, 1\times 10^{-1}]$            & $5\times 10^{-3}$               & \\ [2pt] \cline{1-4}
\end{tabular}
\end{table}

\begin{table}[!h]
\centering
\caption{Step Size $\delta_{\text{KL}}$ for TNPG}
\label{hyper_npg}
\begin{tabular}{l|lll}
\cline{1-4} \\ [-8pt]
 & Search Range              & Best                 &  \\ [2pt] \cline{1-4} \\ [-8pt]
Cart-Pole Swing Up     & $[1\times 10^{-3}, 5 \times 10^{0}]$            & $5 \times 10^{-2}$               &  \\
Double Inverted Pendulum     & $[1\times 10^{-3}, 5 \times 10^0]$     & $3 \times 10^{-2}$               &  \\
Swimmer               & $[1\times 10^{-3}, 5 \times 10^{0}]$            & $1\times 10^{-1}$               &  \\
Ant                   & $[1\times 10^{-3}, 5 \times 10^{0}]$            & $3 \times 10^{-1}$               &  \\ [2pt] \cline{1-4}
\end{tabular}
\end{table}

\begin{table}[!h]
\centering
\caption{Step Size $\delta_{\text{KL}}$ for TRPO}
\label{hyper_trpo}
\begin{tabular}{l|lll}
\cline{1-4} \\ [-8pt]
 & Search Range              & Best                 &  \\ [2pt] \cline{1-4} \\ [-8pt]
Cart-Pole Swing Up     & $[1\times 10^{-3}, 5 \times 10^{0}]$            & $5 \times 10^{-2}$               &  \\
Double Inverted Pendulum     & $[1\times 10^{-3}, 5 \times 10^{0}]$     & $1 \times 10^{-3}$               &  \\
Swimmer               & $[1\times 10^{-3}, 5 \times 10^{0}]$            & $5 \times 10^{-2}$               &  \\
Ant                   & $[1\times 10^{-3}, 5 \times 10^{0}]$            & $8 \times 10^{-2}$               &  \\ [2pt] \cline{1-4}
\end{tabular}
\end{table}

\begin{table}[!h]
\centering
\caption{Step Size $\delta_{\text{KL}}$ for REPS}
\label{hyper_reps}
\begin{tabular}{l|lll}
\cline{1-4} \\ [-8pt]
 & Search Range              & Best                 &  \\ [2pt] \cline{1-4} \\ [-8pt]
Cart-Pole Swing Up     & $[1\times 10^{-3}, 5\times 10^0]$            & $1\times 10^{-2}$               &  \\
Double Inverted Pendulum     & $[1\times 10^{-3}, 5\times 10^0]$     & $8 \times 10^{-1}$               &  \\
Swimmer               & $[1\times 10^{-3}, 5\times 10^0]$            & $3 \times 10^{-1}$               &  \\
Ant                   & $[1\times 10^{-3}, 5\times 10^0]$            & $8 \times 10^{-1}$               &  \\ [2pt] \cline{1-4}
\end{tabular}
\end{table}

\begin{table}[!h]
\centering
\caption{Initial Extra Noise for CEM}
\label{hyper_cem}
\begin{tabular}{l|lll}
\cline{1-4} \\ [-8pt]
  & Search Range              & Best                 &  \\ [2pt] \cline{1-4} \\ [-8pt]
Cart-Pole Swing Up     & $[1\times 10^{-3}, 1]$            & $1 \times 10^{-2}$               &  \\
Double Inverted Pendulum     & $[1\times 10^{-3}, 1]$     & $1 \times 10^{-1}$               &  \\
Swimmer               & $[1\times 10^{-3}, 1]$            & $1 \times 10^{-1}$               &  \\
Ant                   & $[1\times 10^{-3}, 1]$            & $1 \times 10^{-1}$               &  \\ [2pt] \cline{1-4}
\end{tabular}
\end{table}

\begin{table}[!t]
\centering
\caption{Initial Standard Deviation for CMA-ES}
\label{hyper_cma_es}
\begin{tabular}{l|lll}
\cline{1-4} \\ [-8pt]
 & Search Range              & Best                 &  \\ [2pt] \cline{1-4} \\ [-8pt]
Cart-Pole Swing Up     & $[1\times 10^{-3}, 1\times 10^3]$            & $1 \times 10^{3}$               &  \\
Double Inverted Pendulum     & $[1\times 10^{-3}, 1\times 10^3]$     & $3 \times 10^{-1}$               &  \\
Swimmer               & $[1\times 10^{-3}, 1\times 10^3]$            & $1 \times 10^{-1}$               &  \\
Ant                   & $[1\times 10^{-3}, 1\times 10^3]$               & $1 \times 10^{-1}$               &  \\ [2pt] \cline{1-4}
\end{tabular}
\end{table}